\documentclass[
twocolumn,
]{ceurart}

\usepackage{listings}
\usepackage{algorithm}
\usepackage{algorithmic}

\usepackage{booktabs}
\usepackage{latexsym}
\usepackage{amsmath}
\usepackage{hyperref}
\usepackage{multirow}
\usepackage{xcolor}
\newcommand{\dotr}{\mbox{$\boldsymbol{\cdot}$}}
\definecolor{myOrange}{HTML}{F8CBAD} 
\newcommand{\norm}[1]{\left\lVert#1\right\rVert}
\newcommand{\xdownarrow}[1]{%
  {\left\downarrow\vbox to #1{}\right.\kern-\nulldelimiterspace}
}

\begin{document}

\copyrightyear{2024}
\copyrightclause{Copyright for this paper by its authors.
  Use permitted under Creative Commons License Attribution 4.0
  International (CC BY 4.0).}

\conference{Machine Learning for Cognitive and Mental Health Workshop (ML4CMH), AAAI 2024, Vancouver, BC, Canada.}

\title{Learning to Generate Context-Sensitive Backchannel Smiles for Embodied AI Agents with Applications in Mental Health Dialogues}


\author[1]{Maneesh Bilalpur}[%
email=mab623@pitt.edu,
]
\cormark[1]
\address[1]{University of Pittsburgh, Pittsburgh, Pennsylvania, USA}

\author[2]{Mert Inan}[%
email=inan.m@northeastern.edu,
]
\address[2]{Northeastern University, Boston, Massachusetts, USA}

\author[2]{Dorsa Zeinali}[%
email=zeinali.d@northeastern.edu,
]

\author[1]{Jeffrey F. Cohn}[%
email=jeffcohn@pitt.edu,
]

\author[2]{Malihe Alikhani}[%
email=m.alikhani@northeastern.edu,
]

\cortext[1]{Corresponding author.}

\maketitle

\begin{abstract}
Addressing the critical shortage of mental health resources for effective screening, diagnosis, and treatment remains a significant challenge. This scarcity underscores the need for innovative solutions, particularly in enhancing the accessibility and efficacy of therapeutic support. Embodied agents with advanced interactive capabilities emerge as a promising and cost-effective supplement to traditional caregiving methods. Crucial to these agents' effectiveness is their ability to simulate non-verbal behaviors, like backchannels, that are pivotal in establishing rapport and understanding in therapeutic contexts but remain under-explored. To improve the rapport-building capabilities of embodied agents we annotated backchannel smiles in videos of intimate face-to-face conversations over topics such as mental health, illness, and relationships. We hypothesized that both speaker and listener behaviors affect the duration and intensity of backchannel smiles. Using cues from speech prosody and language along with the demographics of the speaker and listener, we found them to contain significant predictors of the intensity of backchannel smiles. Based on our findings, we introduce backchannel smile production in embodied agents as a generation problem. Our attention-based generative model suggests that listener information offers performance improvements over the baseline speaker-centric generation approach. Conditioned generation using the significant predictors of smile intensity provides statistically significant improvements in empirical measures of generation quality. Our user study by transferring generated smiles to an embodied agent suggests that agent with backchannel smiles is perceived to be more human-like and is an attractive alternative for non-personal conversations over agent without backchannel smiles.
\end{abstract}


\section{Introduction}
Fewer 
than a third of the US population has sufficient access to mental health professionals \cite{modi2022exploring}.
This highlights the need for additional resources to help mental health professionals meet the community's demands. Problems like symptom detection and evaluating treatment efficacy have made great strides with AI \cite{song2020spectral, ceccarelli2022multimodal, yang2012detecting}. Alternatively, Embodied agents such as Ellie \cite{devault2014simsensei} through their multimodal behavior offer promising solutions to support the growing mental health needs. However, the development of such systems presents numerous challenges. These include the scarcity of mental health-related datasets, limited access to domain experts for designing reliable and robust systems, and the ethical considerations crucial to their design and adaptation. 
\begin{figure}[!t]
    \centering
    \includegraphics[width=\linewidth]{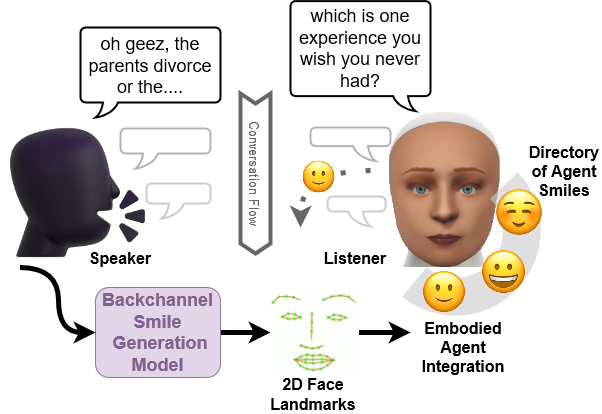}
    \caption{Overview of steps for backchannel smile generation in an embodied agent in a human-agent interaction: Speaker and listener (agent) turns are used to generate the listener's response facial expression as landmarks. The landmarks are then integrated with the embodied agent and added to the conversation flow represented as a dotted arrow.}
    \label{fig:system_schematic}
\end{figure}
Among such challenges, one aspect that stands out is the agent's ability to establish a common ground with users. Addressing this is particularly crucial when the agent functions as a listener. Effective grounding in such scenarios relies heavily on multimodal non-verbal behaviors like backchannels. These subtle yet impactful cues are pivotal in building rapport and understanding between the user and the agent. Hence, understanding and incorporating these behaviors into embodied agents is not only challenging but also essential for creating a supportive and empathetic environment for individuals seeking mental health support. Addressing these challenges can pave the way for more effective, accessible, and empathetic digital mental health interventions.

In dyadic conversations, at any given time one person may have the floor (i.e., is speaking) while the other is listening. Backchannels (BC) refer to behaviors of the listener that do not interrupt the speaker. BCs signal attention, agreement, and emotional response to what is said. Inappropriate BC smiles such as ones that appear too short or too long or for which the timing appears ``off'' can disrupt the conversational rapport and result in unsuccessful or disrupted conversations. Our objective is to understand appropriate BC smiles from dyadic conversations and how an embodied agent can employ them when interacting with a human.

 
 Conversational agents typically realize BC smiles using rule-based systems, discriminative approaches, or sometimes simply mimicking the smiles of the speaker. Mimicking, however, fails to generalize to situations that require a contextually relevant smile. And rule-based and discriminative approaches offer limited coverage due to the diversity of smiles \cite{ambadar2009all}.




We present a generative approach for BC smiles in listeners to address these limitations and enable contextually relevant BC smiles in embodied agents. An overview of the approach is presented in Figure~\ref{fig:system_schematic}. Unlike existing works that solely depend on speaker behavior for BC production (see related work section), we use both speaker and listener behaviors to study how they affect the intensity and duration of the BC smile. We use cues from prosody, language, and the demographics of dyads to identify statistically significant predictors (referred to as a conditioning vector) of smiles. In addition to the audio features from both interaction participants, we leverage the conditioning vector in generating the BC smiles. In this paper, we:  

\begin{enumerate}
    \item Annotate backchannel smiles in a face-to-face interaction dataset\footnote{Data and code: \href{https://github.com/bmaneesh/Generating-Context-Sensitive-Backchannel-Smiles/}{https://github.com/bmaneesh/Generating-Context-Sensitive-Backchannel-Smiles/}} 
    of dyads that differ in their composition of biological sex and type of relationship. 

    \item Present our statistical analysis to identify various speaker and listener-specific cues that significantly predict the duration and intensity of backchannel smiles.
    
    \item Generate backchannel smiles using an attention-based generative model that uses the listener and speaker turn features with the identified significant predictors.

    \item Bridge the gap between the model-based generation of non-verbal behaviors (as facial landmarks) and their physical realization by emulating the generated behavior with an embodied agent.

    \item Show that our BC smile generation yields appropriate and natural-looking smiles through a user study involving the embodied agent. 
\end{enumerate}

Results suggest speaker sex, their use of negations, loudness, word count in the listener's turn, their usage of comparisons, and mean pitch are significant predictors of BC smile intensity. Our generative approach shows that taking listeners' behavior into account improves performance, and adding the conditioning vector offers significant improvements in terms of empirical metrics such as Average Pose Error (APE) and Probability of Correct Keypoints (PCK).

\section{Related Work}


Existing works have validated the efficacy of an agent-driven conversation in mental health dialogue and counseling situations. \citet{devault2014simsensei}, through their agent-based interviews for distress and trauma symptoms, found that participants were comfortable interacting with the agent as well as sharing intimate information. \citet{utami2019collaborative} used embodied agents for couples counseling. Participants reported significantly improved affect and intimacy with their partner and generally enjoyed the agent-driven counseling session. Our work builds on this line of research to improve the BC capabilities of agents.



Backchannel behaviors were traditionally produced using a set of predefined rules based on prosodic or linguistic cues of the speaker. Both \citet{ward2000prosodic, benus2007prosody} have found prosodic cues (particularly pitch and its changes) to be reliable predictors for vocal BC occurrence. In contrast, we use prosody and linguistic cues from both speaker and listener to identify significant predictors of BC smiles. 


In the multimodal context, \citet{bertrand2007backchannels} studied prosodic, morphological, and discourse markers for their effect on vocal and gestural backchannels (hand gestures, smiles, eyebrows), and \citet{truong2011multimodal} 
explored visual BCs by often limiting them to head nods and, at times, grouping different BCs into the same category \cite{gravano2009backchannel} without accounting for their intrinsic differences. They depended on the speaker's behavior to identify the occurrence and ignored the listener. In addition to leveraging the listener behavior, we specifically study smiles because of their diversity and include both unimodal (visual) and bimodal (visual together with vocal activity) BC smiles.

\citet{wang2018every} introduced diversity in generated smiles by conditioning on a specific class and sampling using a variational autoencoder. Learn2Smile \cite{feng2017learn2smile} used the facial landmarks of the speaker to generate complete listener behavior by separately predicting the low-frequency (nods) and high-frequency (blinks) components of facial motion. \citet{ng2022learning} leverage the speaker and listener's motion and speech features to predict the listener's future motion information. Unlike earlier works that have been limited to facial expression generation using landmarks, their usage of 3D Morphable Models to define facial expressions offers a flexible solution to generate realistic facial expressions in the presence of diverse head orientations. These solutions focus on the entire listener's behavior and offer no insights about specific BC behaviors. Their integrations are also limited to 3D Morphable Models.


The BC smiles produced in this work not only leverage the speaker and listener activity but also condition the generation on salient factors that were found to be significant predictors of smile attributes -- duration (the time elapsed between the onset of a smile and its offset) and intensity (maximum amplitude of a smile). Using an embodied agent, we also bridge the gap between generated landmarks and their physical realization. 

\section{Dataset}

One of the primary challenges in studying non-verbal behavior in mental health interactions is access to an appropriate dataset. Patient-therapist interactions or interactions with mental health professionals are access-restricted to protect the identifiable information of the individuals. As a result, we use a YouTube-based large-scale dataset of face-to-face dyadic interactions--RealTalk \cite{geng2023affective}. The RealTalk dataset consists of individuals taking turns asking predefined, intimate questions about family, dreams, relationships, illness, and mental health\footnote{The original videos can be accessed from \url{https://www.youtube.com/c/TheSkinDeep}}. We believe intimate conversations are among the closest accessible alternatives to studying BC behaviors for mental health applications. In this section, we elaborate on our contributions in terms of the annotations for BC smiles and discuss how they differ by the demographics of the dyads and features from the speaker and listener turn preceding it. 




\begin{figure}[!h]
    \centering
    \includegraphics[width=.7\columnwidth]{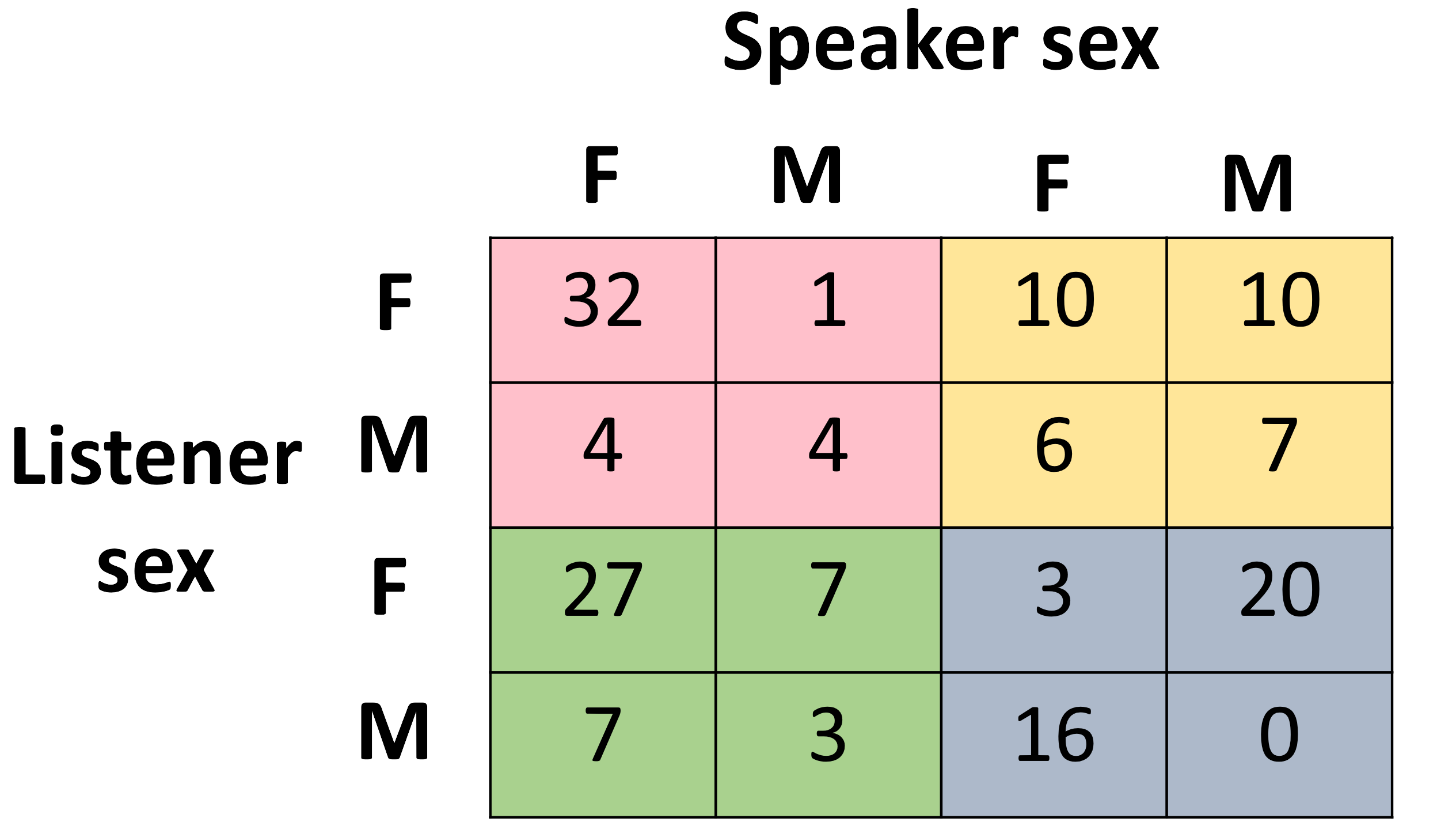}
    \caption{Distribution of speaker and listener sex across different interpersonal relationships in annotated RealTalk dataset. Relationships are color-coded: siblings (pink), friends (orange), paternal (green), and romantic couple (grey).}
    \label{fig:gender-relationship distribution}
\end{figure}


\subsection{Annotating Backchannel Smiles}
We manually annotated 191 BC smiles from 48 (out of 692) dyadic interactions in the RealTalk dataset. The dyads comprised male and female participants from different ethnicities, and social relationships such as siblings, paternal, romantic, and fraternal. The smiles were nearly balanced across the different interpersonal relationships (see Figure~\ref{fig:gender-relationship distribution}). An automated facial expression prediction framework \cite{ertugrul2019afar} was used to evaluate the reliability of the manual annotations. About 83\% (i.e., 158 smiles) of the 191 annotated smiles had an A-level or higher intensity. One outlier smile was dropped because of the extremely long duration. The resultant 157 smiles, along with their predicted intensity, were used in this work. In addition to the video recordings at 25 fps and 720p resolution, the dataset also contains speaker-identified turn-level text obtained through automatic transcription \cite{schneider2019wav2vec}. The individuals in the dyadic interaction occupied fixed positions (left and right) in the videos. In this work, the biological sex of the participants was inferred from the videos. Videos where sex could not be established with confidence were discarded.



\subsection{Effect of Sex and Relationship on Smile Attributes}
Given various interpersonal relationships in the dataset of individuals of both sexes, we compared the mean duration of backchannel smiles across the factors using ANOVA (Table~\ref{tab:anova_duration}) with type-III sum of squares to account for imbalance between males and females. Two-way interactions between sex, and sex and relationship were also included. The ANOVA analysis suggests that the duration of backchannel smiles differs significantly by listener sex and the interaction effect of the listener sex and relationship. A post hoc Tukey revealed that male listeners, when interacting with their siblings (regardless of speaker sex), express longer BC smiles (p\textless0.05).

\begin{table}[!h]
\centering
\caption{ANOVA of listener sex, speaker sex, and relationship on duration of smile. `*' indicates p\textless0.05 and `**' indicates p\textless0.01).}
\resizebox{\columnwidth}{!}{
\begin{tabular}{lrrrrr}
  \toprule
 & Df & Sum Sq & Mean Sq & F value & Pr($>$F) \\ 
  \midrule
$sex_{listener}$                & 1 & 12.36 & 12.36 & 4.59 & 0.0339 * \\ 
  $sex_{speaker}$                 & 1 & 1.29 & 1.29 & 0.48 & 0.4907 \\ 
    $relationship$                   & 3 & 4.18 & 1.39 & 0.52 & 0.6709 \\ 
    \begin{tabular}{c}
         $sex_{listener}*$ \\
         $relationship$
    \end{tabular}  & 3 & 42.80 & 14.27 & 5.29 & 0.0017 ** \\ 
    \begin{tabular}{c}
         $sex_{listener}*$ \\
         $sex_{speaker}$
    \end{tabular} & 1 & 0.90 & 0.90 & 0.33 & 0.5652 \\ 
    \begin{tabular}{c}
         $sex_{speaker}*$ \\
         $relationship$
    \end{tabular} & 3 & 9.70 & 3.23 & 1.20 & 0.3123 \\ 
    \midrule
  Residuals                      & 144 & 388.03 & 2.69 &  &  \\ 
   \bottomrule
\end{tabular}
}
\label{tab:anova_duration}
\end{table}




Similarly, the intensity of smiles marginally differed by the speaker's sex. The post hoc Tukey revealed that the smiles as a response to a male speaker are less intense than a female speaker (p\textless0.1). ANOVA analysis is presented in the appendix as Table \ref{anova_intensity}.

\subsection{Effect of Context Cues}
\label{context_cues}

Our contextual cues were extracted from prosody and speech features independently derived from the turns of both the speaker and the listener just before the smile onset. Since the speaker's turn continues while the listener backchannels, speaker activity till the onset of the smiles was considered in this study. The audio was trimmed to the onset to obtain corresponding contextual cues, and the Montreal Forced Aligner (MFA) \cite{mcauliffe17_interspeech} was used to extract corresponding transcription information.

\begin{figure}[h]
    \centering
    \includegraphics[width=\linewidth]{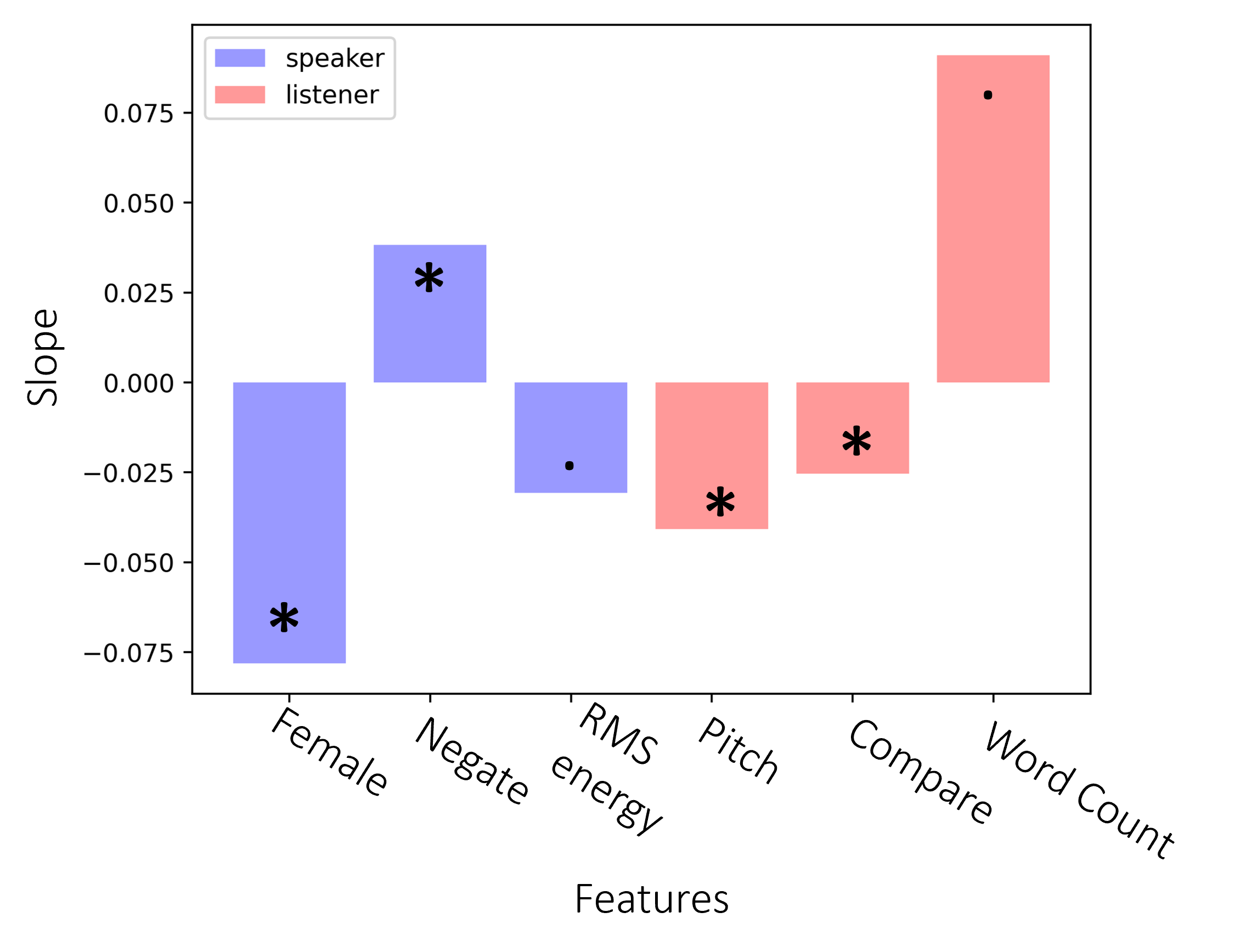}
    \caption{Regression slopes showing the effect of context cues on the intensity of BC smiles. A positive slope indicates the smile intensity increases with a given feature (vice-versa for a negative slope). * indicates slope is significant at p\textless0.05 and ${\dotr}$ indicates marginal significance at p\textless0.1.} 
    \label{fig:intensity_slopes}
\end{figure}

\paragraph{Prosody cues:} Our prosodic features consisted of some of the fundamental characteristics of speech, such as mean pitch during the turn, range of the pitch, and Root Mean Square (RMS) energy of the audio signal. These features were chosen because of their relevance (see related work) in BC behavior and also due to the ease of interpretation as well as their ability to convey various behavioral traits. For example, RMS energy conveys traits such as confidence, doubtfulness, and enthusiasm \cite{memon2020acoustic}. Lastly, using the OpenSMILE \cite{eyben2010opensmile} software, prosodic features were obtained.

\paragraph{Speech cues:} The spoken content of speaker and listener turns was also accounted for through variables from the Linguistic Inquiry and Word Count (LIWC) \cite{pennebaker2015development} framework. These variables were word count, usage of negations (no, not, never), comparisons (greater, best, after), interrogative words (how, when, what), valence of the turns (positive or negative emotion), and focus on events in the past, present and future. 


A generalized linear model predicted the smile intensity from context cues and dyad demographics. Results using an inverse link function (model explained variance $R^{2}=0.243$) with the prosody and speech cues from the audio signal are presented as Figure~\ref{fig:intensity_slopes}. Note that the speakers' and listeners' context cues were Z-score normalized. Speaker characteristics such as sex and negations were found to be significant predictors of intensity. Female speakers elicited significantly narrower smiles from their listeners, but the speaker's usage of negations resulted in wider smiles. The speaker's loudness (RMS energy) had a marginally significant negative correlation with the smile intensity. Listener behavior also significantly impacted their BC smiles. Using comparative words by the listener and their mean pitch in their preceding turn resulted in significantly narrower smiles. In contrast, their word count had a marginally significant positive correlation with intensity. A similar analysis for duration did not reveal any significant correlations.

\section{Modeling Smiles}

To automatically generate BC smile and non-smile activity in listeners, we use the audio from the speaker's current turn and the listener's last turn as input. 15 smiles were dropped due to difficulties in the preprocessing steps with MFA. The remaining 142 annotated smile instances were augmented with an equal number of non-smile instances. The non-smile instances were identified so that they were at least two seconds away from the onset of the closest smile instance, a strategy adopted from \cite{ekstedt2022voice} for turn-taking prediction. The mean duration of smiling and non-smiling instances was ensured to be the same.

\begin{figure*}[!ht]
    \centering
    \includegraphics[width=\textwidth]{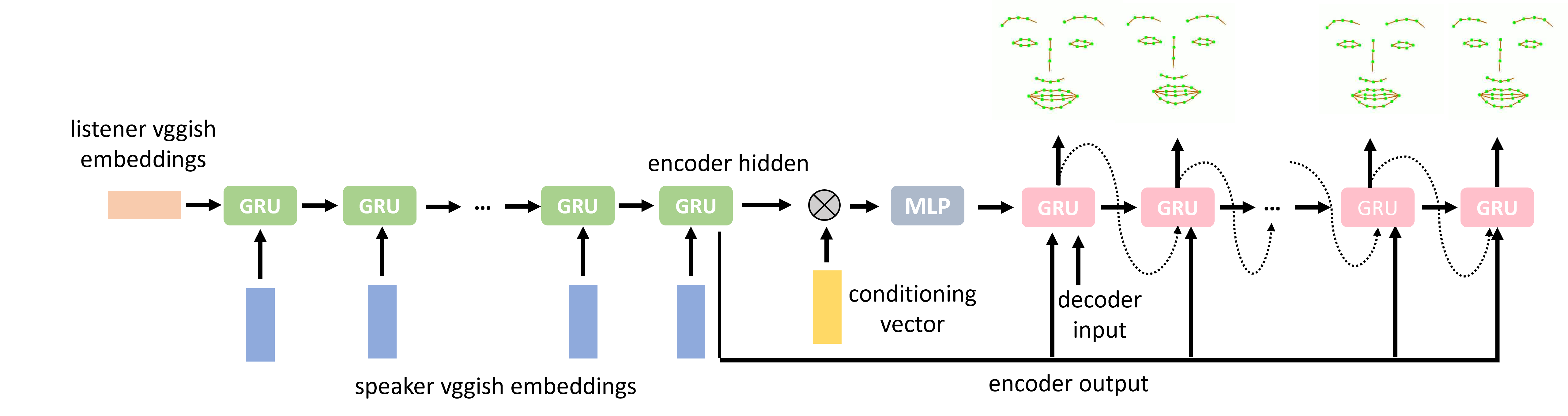}
    \caption{Architecture of a generative model incorporating the significant predictors (conditioning vector) for backchannel smiles. Encoder input contains speech embeddings of listener and speaker from the pretrained vggish model. The encoder's final hidden state is concatenated with the conditioning vector and then used to initialize the decoder's hidden state. Decoder output landmarks are sequentially fed (dotted curves) to generate the next landmarks in the output sequence.} 
    \label{fig:generative_model}
\end{figure*}

\paragraph{Attention-based generative model:} The generative model (Figure~\ref{fig:generative_model}) for facial landmark prediction primarily consisted of an encoder and a decoder with a one-layer GRU each. Inputs to the model were embeddings from speaker and listener turns extracted using the pretrained vggish model \cite{hershey2017cnn}. We limited the input context length to use turn durations of 60 seconds. The output context was limited to predicting one second of facial activity. The speaker vggish embeddings were used as input to the encoder. The hidden state of the GRU was initialized as the mean of the listener's turn embeddings. The final hidden state of the encoder was concatenated with the conditioning vector, and a linear layer with ReLU activation was used to match the dimensionality of the decoder's hidden state. At each decoding step, attention \cite{bahdanau2014neural} was applied between the encoder output and the decoder's last hidden state (Equation~\ref{eqn:attn}) to use as the input to the next step.


\begin{equation}
    a(s_{t-1}, h_{i}) = v^{T}tanh(W_{a}h_{i} + W_{b}s_{t-1})
    \label{eqn:attn}
\end{equation}

where $a(s_{t-1}, h_{i})$ is the attention between decoder last hidden state ($s_{t-1}$) and encoder output ($h_{i}$). $W_{i}$s and $v$ are linear layers.


\subsection{Implementation details}
The videos were split into two vertical halves, one corresponding to each individual in the dyadic interaction. These were used for facial landmark extraction using the AFARtoolbox \cite{ertugrul2019afar}. To account for various facial shapes, we normalized landmarks to the mean face of the dataset using the approach described in \cite{stoll2018sign}. Because of the high degree of correlation between successive frames, frames were downsampled by a factor of three, to use every third frame. Displacement was then calculated as the difference between the landmarks from successive frames. These were further subjected to a min-max normalization to allow for individual differences in smiling dynamics. The normalized displacements were predicted using the attention-based generative model. The predicted frame-level displacements were incorporated into the last known listener facial expression to generate the sequence of facial landmarks recursively. 

We enforced teacher-forcing with simulated annealing during training and linearly decreased the likelihood of using ground truth at every 20 epochs. Stochastic Gradient Descent with a learning rate initialized at $1e-4$ weight decay and $0.99$ momentum were used to minimize the Mean Squared Error (MSE) between predictions and the ground truth. The learning rate was halved when validation loss plateaued for 20 consecutive epochs. Data was partitioned into 75 (train), 15 (validation), and 15 (test) split in terms of the number of dyads. Models were trained for 250 epochs, and validation loss was used to determine the best model for testing. This was repeated 10 times to evaluate the statistical significance of differences against baseline speaker-based BC generation setting. 



\paragraph{Metrics:} Objective measures of performance from gesture generation approaches, including Average Pose Error (APE) and Probability of Correct Keypoints (PCK), were adopted to quantify the generated landmarks against the ground truth from the AFAR toolbox. APE (Equation \ref{eqn:ape}) is equivalent to the mean squared error between predicted facial expression and ground truth facial expression. PCK (Equation \ref{eqn:pck}) is a proximity-based metric that considers the landmark to be correctly predicted if the difference with ground truth falls below a margin. We report mean PCK for $\sigma=0.1$ and $0.2$.


\begin{equation}
    APE = \frac{1}{k} \sum_{y=1}^k \norm{ (\hat{y}(p) - y(p)) }_{2}
    \label{eqn:ape}
\end{equation}

where $k$ is the number of landmarks, $\hat{y}(p)$ is the prediction and $y(p)$ is the groundtruth.

\begin{equation}
    PCK_{\sigma} = \frac{1}{k} \sum_{y=1}^k \delta(\norm{ (\hat{y}(p) - y(p)) }_{2} \leq \sigma)
    \label{eqn:pck}
\end{equation}

where $\delta$ is an indicator function and $\sigma$ is the margin.

\subsection{Results}
\begin{table}[!ht]
\centering
\caption{Average Pose Error (APE) and Probability of Correct Keypoints (PCK) metrics for generated facial expressions under various experimental settings. A downward-facing arrow indicates lower value implies better generation. `*' indicates significance with p \textless 0.05 with `${\dotr}$' indicates marginal significance with p \textless 0.1.}
\begin{tabular}{cccc}
\hline
\textbf{Model}                                                     & \textbf{APE$\downarrow$} & \textbf{PCK$\uparrow$} \\ \hline
Speaker only (Baseline)                            & 9.552 & 0.219 \\
Speaker and Listener                         & 9.346$^{\dotr}$ & 0.220$^{\dotr}$ \\
\begin{tabular}{c}
    Speaker and Listener with \\
    Conditioning vector 
\end{tabular} & 9.279* & 0.223* \\
Speaker and Conditioning vector            & 9.615    & 0.218$^{\dotr}$ \\ \hline



\end{tabular}
\label{tab:generation_results}
\end{table}


Using listener behavior and conditioning vector together with the speaker behavior resulted in improved performance compared to the baseline speaker behavior-based prediction. As shown in Table~\ref{tab:generation_results}, APE decreased by 0.273 points while PCK increased by 0.004; these gains were statistically significant. When listener behavior was added to the speaker behavior, marginally significant improvements were observed. APE reduced by 0.206 points while PCK increased by 0.001 points. These reiterate our hypothesis that both speaker and listener contribute to BC behaviors. When speaker behavior was augmented with the conditioning vector, only nominal differences were observed against the baseline. APE increased by 0.063 points, and PCK decreased by 0.001.

\begin{figure}[!t]
    \centering
    \includegraphics[width=\linewidth]{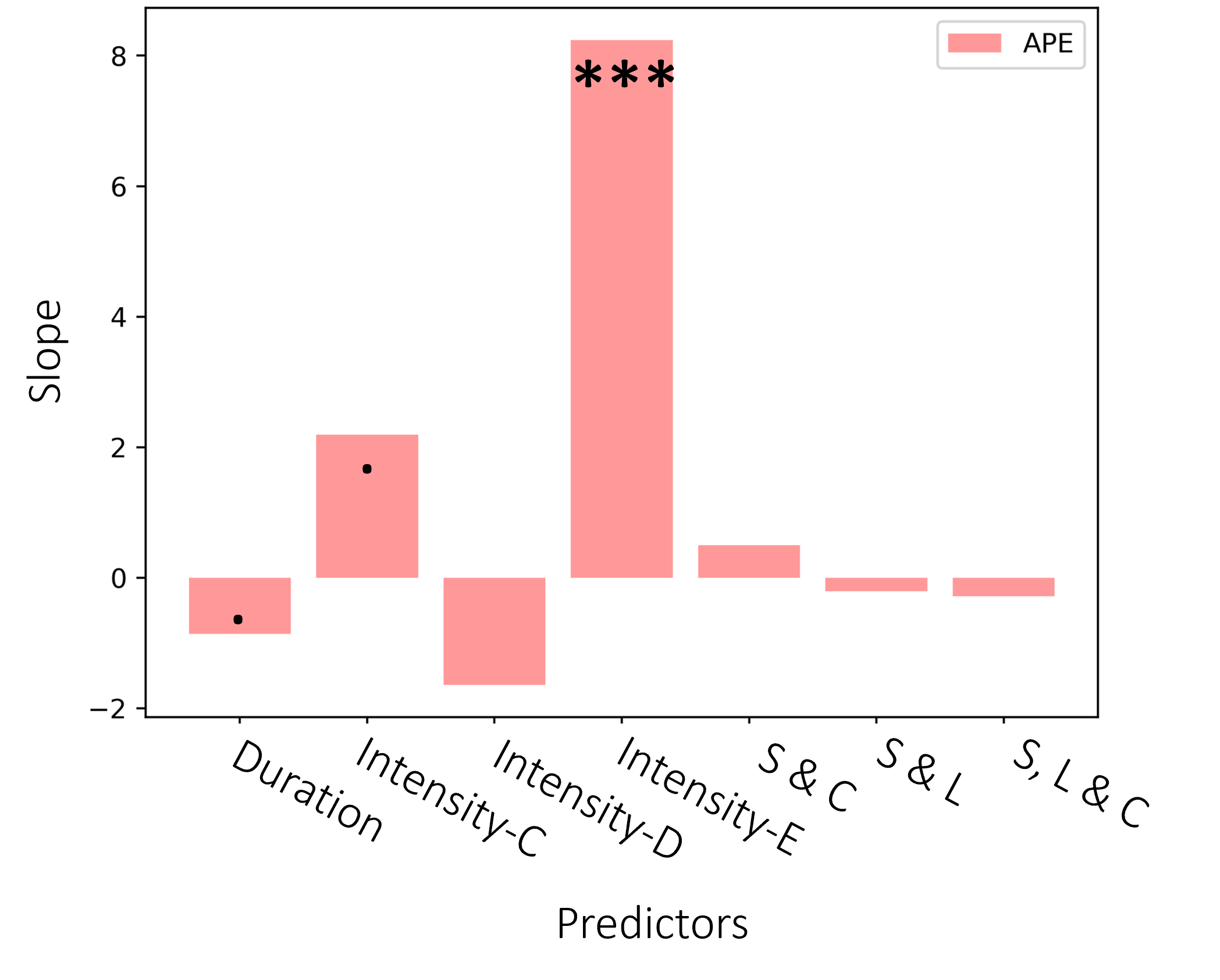}
    \includegraphics[width=\linewidth]{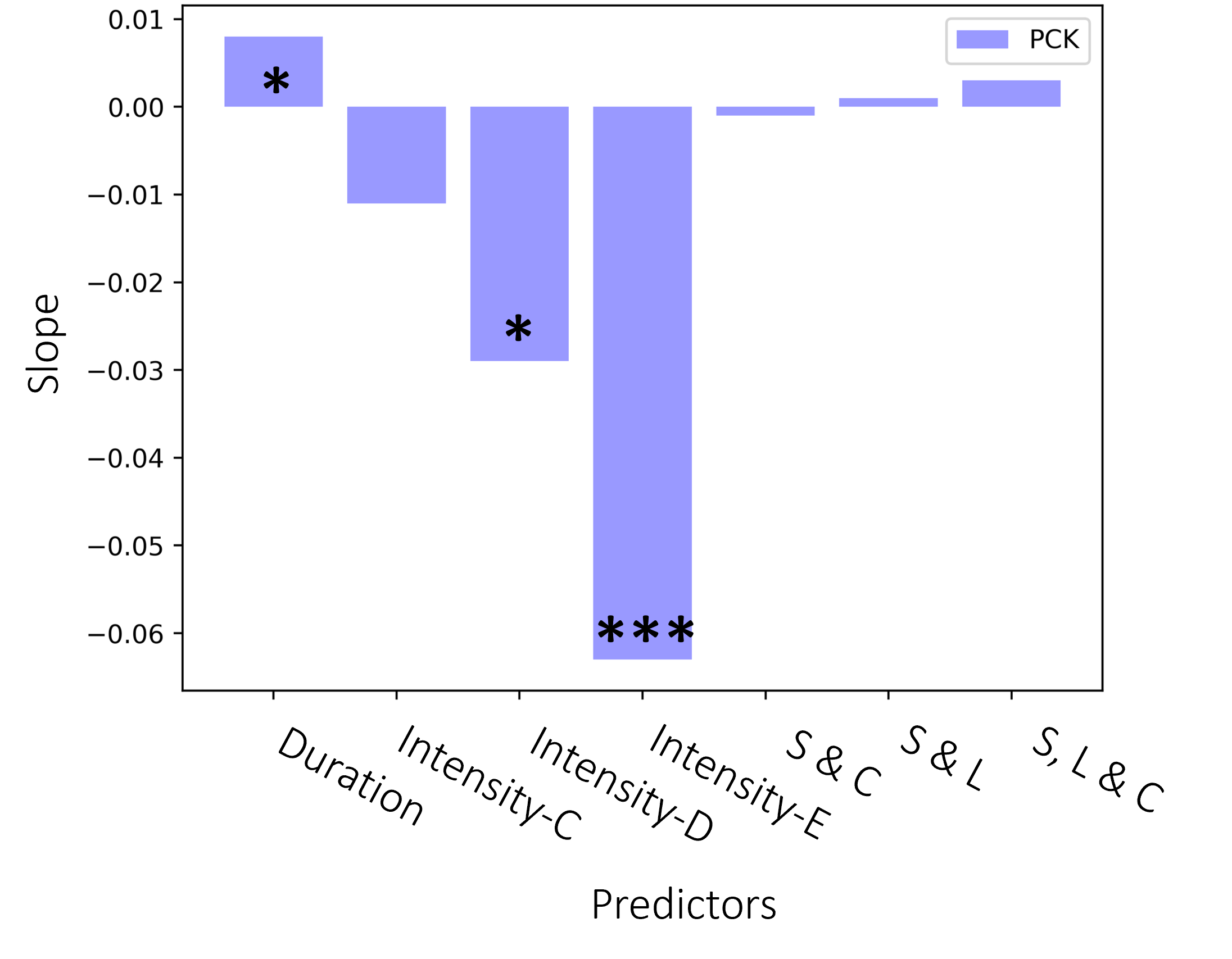}
    \caption{Effect of duration and intensity of smile along with ablation of inputs on generative model performance measured using APE (top) and PCK (bottom). S \& C-speaker and conditioning vector, S \& L-speaker and listener, and S, L \& C-speaker and listener and conditioning vector as inputs to the model. `${\dotr}$', `*' and `***' indicate significance with p \textless 0.1, p \textless 0.05 and p \textless 0.001 respectively. \vspace{-5pt}} 
    \label{fig:effect_on_perf}
\end{figure}


To understand how the performance varies with different smiles, we predicted APE (and PCK) as a linear combination of duration, intensity, and the model configuration using a regression model. Results from Figure~\ref{fig:effect_on_perf} show that duration significantly affects the PCK. Interestingly, the positive slope suggests that longer smiles are generated better over shorter smiles. Only a marginally significant effect of duration can be observed for APE. With the increase in the intensity of the smile, the generation performance decreases. This is significant for D-level and E-level smiles. Using listener features and the conditioning vector along with the speaker features improves the performance (negative and positive slopes for APE and PCK, respectively) compared to the baseline speaker-based generation. However, this effect is not statistically significant.



\begin{figure*}[t]
    \centering
    \includegraphics[width=.8\textwidth]{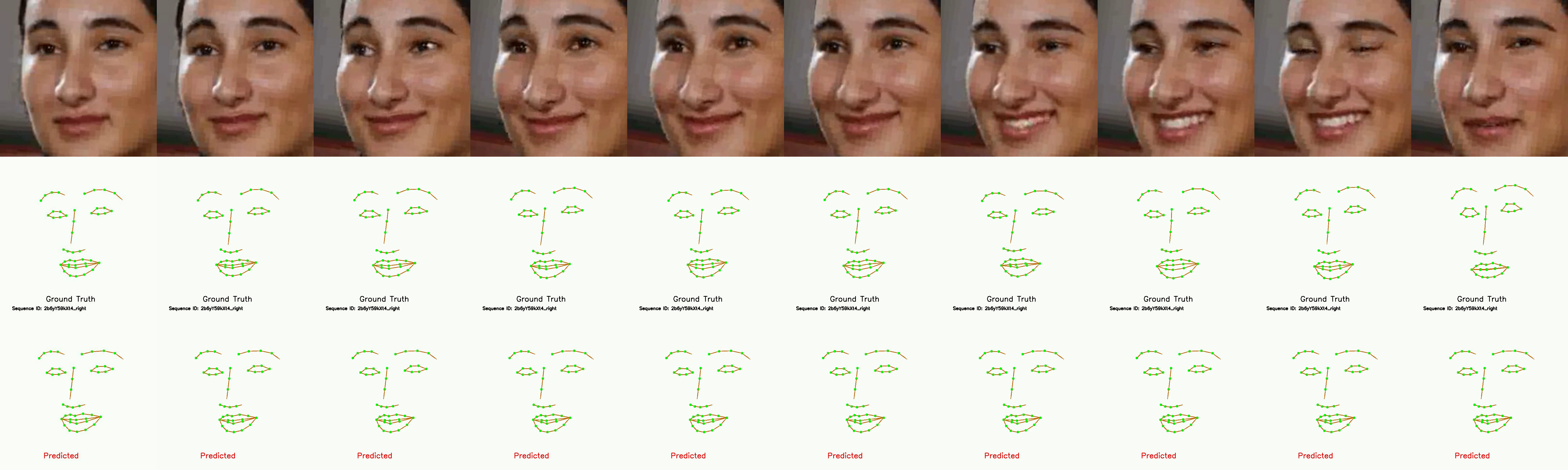}
    \includegraphics[width=.8\textwidth]{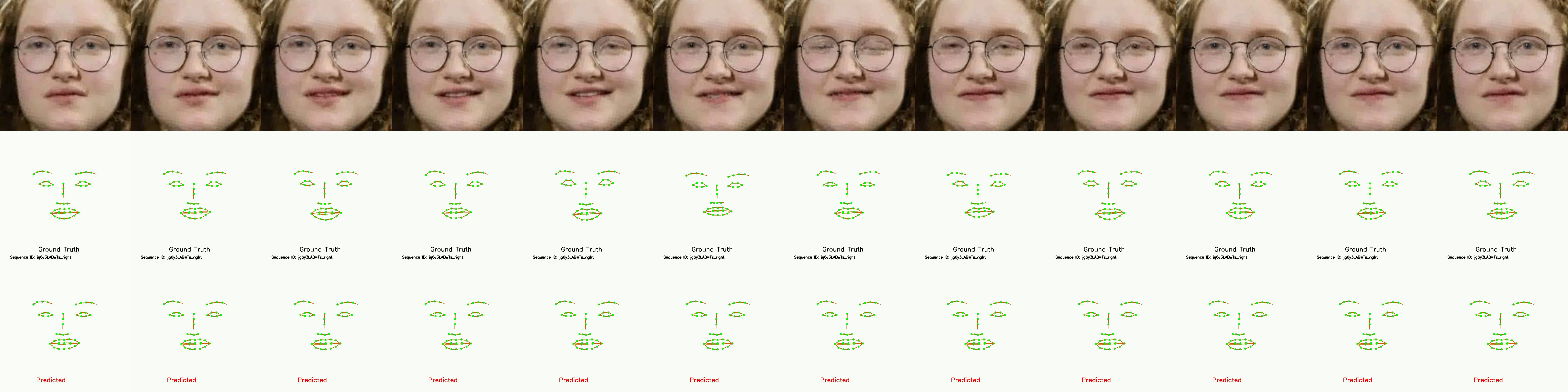}

    \caption{Two sample smiles from the dataset showing their onsets (left-most frame to widest smile frame) and offsets (widest smile frame to right-most frame). Note that while the evolution of smile is noticeable in ground truth landmarks (second row) of the top smile, subtle changes between successive frames of the bottom smile are not captured by its ground truth landmarks. This is also observed in the generated landmarks (third row). Zoom-in recommended. The faces used are from the RealTalk dataset.}
    \label{fig:model_predictions}
\end{figure*}

\begin{figure}[t]
    \centering
    \includegraphics[width=\linewidth]{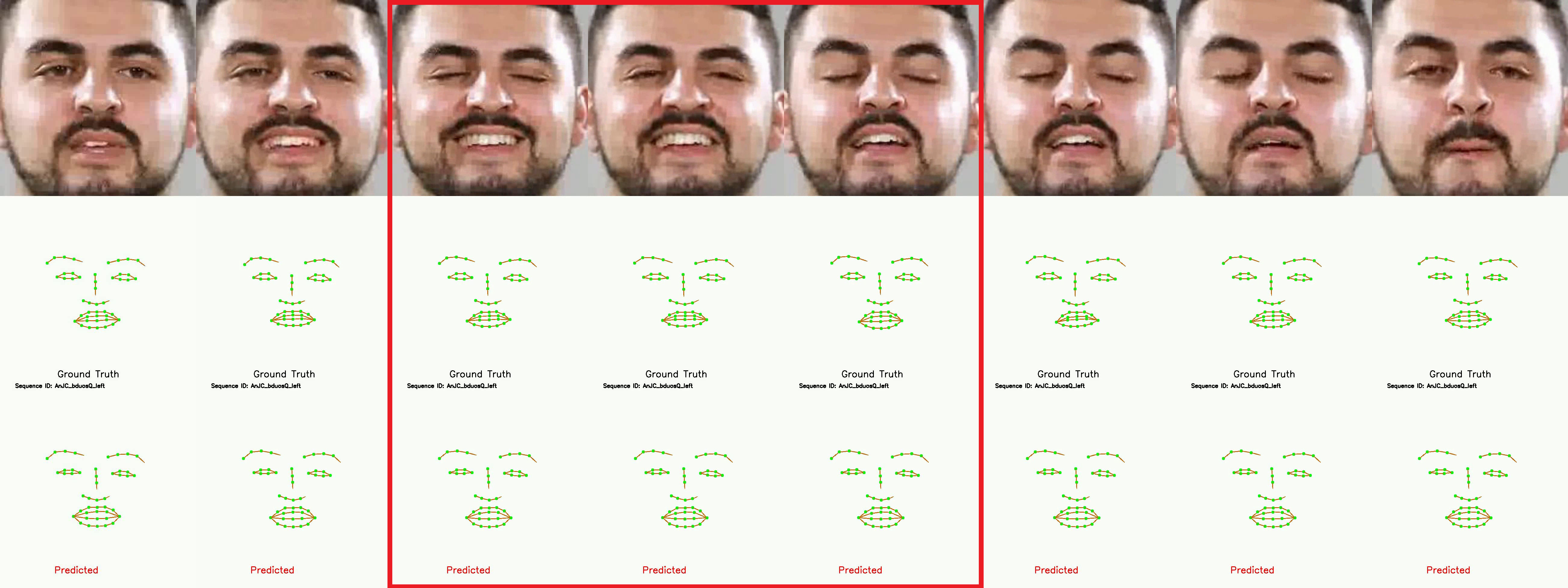}    
    \caption{Limitation of the current approach in generating a bimodal backchannel smile. The frames highlighted in red box correspond to the co-occurring verbal ``yeah''. Notice that ground truth landmarks (second row) fail to capture the vertical mouth movement. This is also observed in the generated landmarks (third row). Zoom-in recommended. The faces used are from the RealTalk dataset.}

    
    \label{fig:vocal_bcs}
\end{figure}

Qualitative evaluation of ground truth landmarks from Figure~\ref{fig:model_predictions} suggest the deficiencies of the existing facial landmark prediction approaches \cite{ertugrul2019afar} to accurately track lip corners both in the presence and absence of non-frontal head pose. While a visually noticeable difference can be observed as the smile evolves, the ground truth landmarks fail to capture the subtle lip corner motion. This limitation in the ground truth has resulted in nominal motion in the predicted landmarks. We also found that BC smiles that co-occur with vocal activity are challenging to predict. Figure \ref{fig:vocal_bcs} shows one example where the vertical distance between the upper and lower lips increases and decreases because of the simultaneous \textit{yeah} utterance. However, the model fails to capture this vertical motion.



Metrics like APE and PCK provide an objective measure of the prediction. However, evaluating concepts such as realism and contextual relevance of the BC prediction requires subjective ratings from human evaluation. A convention in evaluating landmark or keypoint-based generative approaches is the human comparison of predicted keypoints against the ground truth \cite{feng2017learn2smile, ahuja2020no}. While this might work for problems such as gesture generation that involve a strong motion component, evaluating subtle behaviors like facial expressions using a similar strategy could be challenging. To address this concern, we leverage the emulated version of an embodied agent: Furhat \cite{al2012furhat}.

\section{Smiles on an Embodied Agent}
So far, we have shown modeling smiles by generating facial landmarks. However, users in real-world scenarios do not expect to see such abstract representations of faces. Aligning these facial landmarks with embodied agents is key for an interactable conversational agent. To achieve this, we describe the procedure to transfer generated landmarks to an embodied robotic simulation system called Furhat. We then conduct a user study 
for subjective perceived differences in Furhat's behavior due to BC smile.

\subsection{Emulation Setup}
Furhat allows users to control facial expressions using a set of facial parameters called BasicParams\footnote{https://docs.furhat.io/remote-api/\#python-remote-api} (ex. MOUTH\textunderscore SMILE\textunderscore LEFT and MOUTH\textunderscore SMILE\textunderscore RIGHT to control the left and the right lip corners; BROW\textunderscore UP\textunderscore LEFT, BROW\textunderscore UP\textunderscore RIGHT to control the left and right eyebrows, etc.). Our setup uses these parameters to enable the embodied agent's smile and express associated eyebrow actions. The landmarks from a generated smile expression were used to calculate the displacement between successive frames and normalized to the [0, 1] range. For eyebrows, only vertical displacement was used. Our inputs to the Furhat API consisted of the lip corner and eyebrow displacements corresponding to the frame with the widest smile (maximum horizontal displacement between the lip corners). The duration of the Furhat smile was set to the duration of the generated smile. Figure~\ref{fig:furhat_examples} shows an example of the resultant expression. The user study was conducted using the Furhat Desktop SDK. However, we do not foresee difficulties transferring the emulation setup to a physically embodied Furhat.

\subsection{User Study Procedure} 
\begin{figure}[!h]
    \centering
    \includegraphics[width=.7\linewidth]{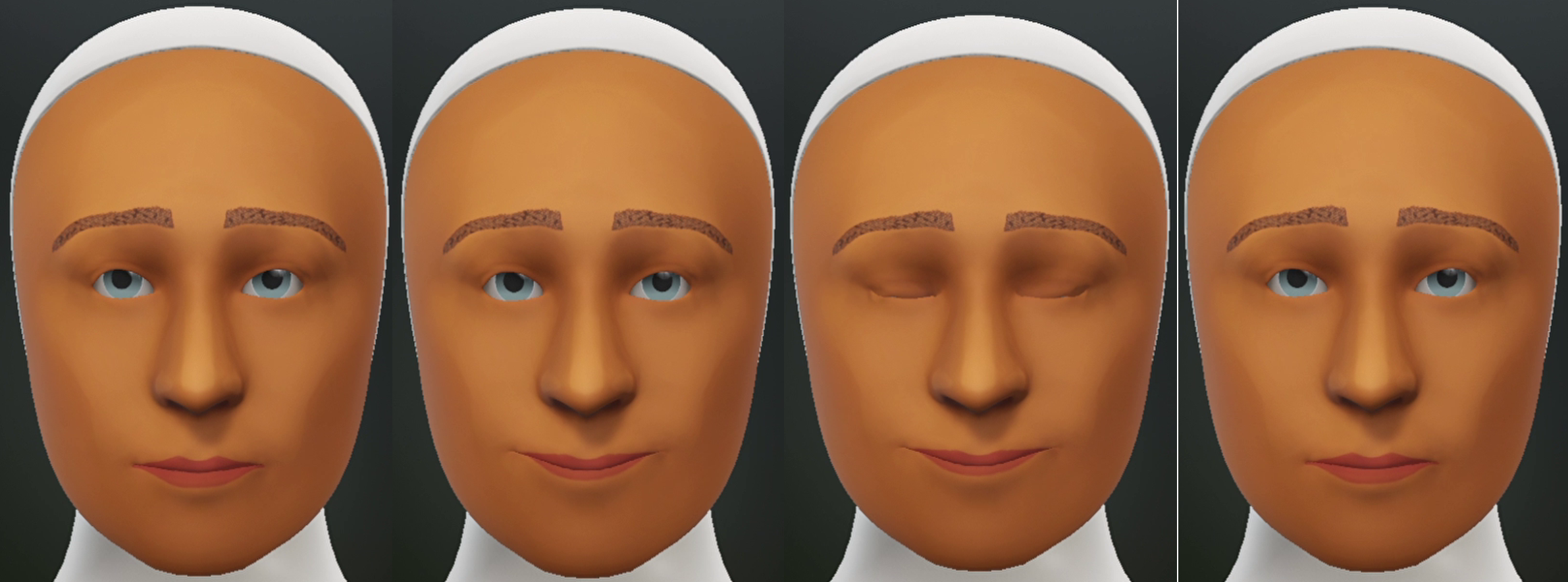}
    \caption{Four frames of an example Furhat robot emulation with different levels of smiles used as backchannels during the conversation in our user study.}
    \label{fig:furhat_examples}
\end{figure}
We conducted a small-scale user study of participants watching two pre-recorded videos of the Furhat interacting with an individual. They differ only in terms of Furhat expressing a BC smile. In both interactions, Furhat starts with a brief introduction of itself, followed by a short question--``How have you been feeling over the last two weeks?''. As the user responds, a smile is generated at the appropriate location (see Figure~\ref{fig:furhat_examples}). We refer to this scenario as the \textit{backchannel} setting. Another video of the same individual interacting with Furhat with no BC (\textit{non-backchannel}) serves as our baseline. Seven graduate students then rated each video recording separately.
Note that raters were not primed on the study's outcome, and no explicit instructions about smiles were given.

To quantify the user's perception of Furhat interacting with an individual, the influence of BC smile in addition to the effect of its intensity and duration, and their willingness to interact with one was quantified through the following questions on a 5-point Likert scale (1: strongly agree, 5: strongly disagree). 

\begin{enumerate}
    \item The Furhat's smiles looked human-like.
    \item The Furhat's smiles looked natural and friendly.
    \item I would talk to this agent frequently.
    \item I felt the brightness of Furhat's smiles was appropriate.
    \item The Furhat was smiling for longer or shorter duration than it was expected.
    \item  I would feel comfortable talking to this agent about non-personal topics.
    \item I would feel comfortable talking to this agent about personal topics.
\end{enumerate}

In addition, open-ended feedback was also a part of the questionnaire. We believe these questions help identify some user-facing challenges in generating BC behaviors and how they influence users' attitudes to embodied agent-based dialogue systems for conversations related to mental health.

\begin{table}[h]
\caption{Number of responses that expressed moderate or strong agreement along various factors related to the BC smiles when interacting with Furhat with and without backchannel behaviors.}
\resizebox{\columnwidth}{!}{%
\begin{tabular}{ccc}
\hline
\textbf{Question}                & \textbf{Backchannel} & \textbf{Non-backchannel} \\ \hline

\textbf{Human-like}              & 5                & 4                    \\
\textbf{Natural}                 & 6                & 6                    \\
\textbf{Willing to interact} & 1                & 0                    \\
\textbf{Appropriate brightness} & 3 & 5 \\
\textbf{Longer or shorter smiles} & 2 & 0 \\
\textbf{Personal conversations}           & 1                & 1                    \\
\textbf{Non-personal conversations}         & 3                & 2                    \\ \hline
\end{tabular}
}
\label{tab:user_study}
\end{table}

\subsection{Results}


Table~\ref{tab:user_study} shows that more users (5/7) expressed moderate or higher agreement that the Furhat agent with BC smile was human-like than its counterpart without BC smile (4/7). One user expressed interest in frequently interacting with the agent in backchannel setting while the lack of backchannels resulted in increased hesitancy among users in frequently using it. Three (out of 7) users found that the brightness of the BC smile was appropriate while two found that the duration of BC smile was longer or shorter than expected. While no difference was observed in terms of users' preference for Furhat for personal conversations based on the presence of the BC smile, more users (3/7) responded that they would use Furhat with BC smiles for non-personal conversations over Furhat without BC smiles (2/7).

\section{Discussion}

Our quantitative results suggest that both speaker and listener behavior are important in generating BC behavior. Using listener behavior together with the conditioning vector offered statistically significant improvements in performance when compared to the baseline speaker-only model. This effect was observed both in terms of APE and PCK. We also found that our attention-based generative model can predict low-intensity smiles better than high-intensity smiles. Our user study shows that more people find our agent human-like when it was able to express BC smiles. Participants prefer to interact with it over the agent with no BC smile capabilities for non-personal conversations. However, for intimate personal conversations, the presence of a BC smile did not sway their decision.

Some limitations of this work include the following. We employed an affordable measure of reliability for BC smile annotations using a prediction model over a human rater. A robust approach would involve at least one more human annotator to perform reliability annotations on a portion of the dataset. The statistical analysis also assumes that the smiles were independent of the individuals and dyads. However, a given individual typically produces multiple smiles. Grouping of smiles by factors such as individuals and dyads can be better modelled using a mixed-effects model. Our user study was designed to demonstrate the feasibility of transferring generated facial landmarks to an embodied agent together with understanding \textit{perceived} differences between interactions with and without BC smiles. An appropriate evaluation framework would include the user interacting with the agent. Followed by a comparison of qualitative subjective ratings of user experience and quantified parameters (such as difference in turn duration, language usage, etc.) of the interaction with and without BC smiles. We believe such approaches provide a holistic evaluation to identify critical instances in the interaction. Lastly, we focused on BC smiles leaving out other conventional signals such as vocal and headpose-based BCs, and how they are affected by the cues from the speaker and listener.


\section{Conclusion}
To enable BCs in embodied agents for mental health applications, we proposed an annotated dataset of face-to-face conversations including topics related to mental health. Our statistical analysis showed that speaker gender together with prosodic and linguistic cues from both speaker and listener turns are significant predictors of the BC smile intensity. Using the significant predictors together with the speaker and listener behaviors to generate BC smiles offers significant improvements in terms of empirical metrics over the baseline speaker-centric generation.

We bridge the gap between conventional non-verbal behavior generation approaches such as landmarks and poses and their realization by showing that generated landmarks can be transferred to an embodied agent. Thus creating the opportunity for evaluation with a human-like manifestation over a traditional evaluation by comparing generated landmark (or keypoint) outputs. Our small-scale user study suggests our Furhat agent that backchannels is more human-like and are more likely to attract users for non-personal interactions. In addition to these contributions, we also discussed some limitations in existing technology towards generating accurate ground truth landmarks through examples such as failure to capture mouth movement in bimodal BCs and how they affect the generated outputs. We believe these limitations also serve as directions for future research. Our work serves as a baseline for computer scientists interested in behavior generation, and an attractive source of BC smiles for behavioral scientists to study the effect of context cues on BC smiles in intimate conversations.



\section{Ethical Statement}
We proposed a generative approach for backchannel smile production to enable naturalistic interactions with embodied AI agents for mental health dialogue. While our dataset offers diverse smiles from people in different interpersonal relationships, like many existing generative approaches, the choice of pretrained embeddings, imbalance between males and females, lack of male-male romantic relationships, and lack of age and ethnicity information in the dataset might have resulted in biased generations. We also acknowledge that using embodied agents in such sensitive applications should undergo rigorous evaluations by technical and domain experts and regulatory bodies. In our work, we do not interpret embodied agents as a substitute for professionals in mental health or allied areas of healthcare but to provide tools for them to better serve the community's demands. We believe that the advantages and limitations of embodied agents in mental health should be presented to the users and the healthcare experts to provide maximum benefits. The information used in this work is identified from a publicly available dataset. Also, special attention has been paid to privacy and copyright requirements for relevant images showing individual faces. The user study raters were voluntary participants, and the University of Pittsburgh IRB approved the data collection.

\section{Acknowledgments}
Bilalpur and Cohn were supported by the U.S. National Institutes of Health through award MH R01-096951. Zeinali was supported through the Khoury Distinguished Fellowship at Northeastern University.

\bibliography{custom}

\begin{thebibliography}{28}
\expandafter\ifx\csname natexlab\endcsname\relax\def\natexlab#1{#1}\fi
\providecommand{\url}[1]{\texttt{#1}}
\providecommand{\href}[2]{#2}
\providecommand{\path}[1]{#1}
\providecommand{\DOIprefix}{doi:}
\providecommand{\ArXivprefix}{arXiv:}
\providecommand{\URLprefix}{URL: }
\providecommand{\Pubmedprefix}{pmid:}
\providecommand{\doi}[1]{\href{http://dx.doi.org/#1}{\path{#1}}}
\providecommand{\Pubmed}[1]{\href{pmid:#1}{\path{#1}}}
\providecommand{\bibinfo}[2]{#2}
\ifx\xfnm\relax \def\xfnm[#1]{\unskip,\space#1}\fi
\bibitem[{Modi et~al.(2022)Modi, Orgera, and Grover}]{modi2022exploring}
\bibinfo{author}{H.~Modi}, \bibinfo{author}{K.~Orgera}, \bibinfo{author}{A.~Grover},
\newblock \bibinfo{title}{Exploring barriers to mental health care in the u.s.}  (\bibinfo{year}{2022}). \DOIprefix\doi{10.15766/rai_a3ewcf9p}.
\bibitem[{Song et~al.(2020)Song, Jaiswal, Shen, and Valstar}]{song2020spectral}
\bibinfo{author}{S.~Song}, \bibinfo{author}{S.~Jaiswal}, \bibinfo{author}{L.~Shen}, \bibinfo{author}{M.~Valstar},
\newblock \bibinfo{title}{Spectral representation of behaviour primitives for depression analysis},
\newblock \bibinfo{journal}{IEEE Transactions on Affective Computing} \bibinfo{volume}{13} (\bibinfo{year}{2020}) \bibinfo{pages}{829--844}.
\bibitem[{Ceccarelli and Mahmoud(2022)}]{ceccarelli2022multimodal}
\bibinfo{author}{F.~Ceccarelli}, \bibinfo{author}{M.~Mahmoud},
\newblock \bibinfo{title}{Multimodal temporal machine learning for bipolar disorder and depression recognition},
\newblock \bibinfo{journal}{Pattern Analysis and Applications} \bibinfo{volume}{25} (\bibinfo{year}{2022}) \bibinfo{pages}{493--504}.
\bibitem[{Yang et~al.(2012)Yang, Fairbairn, and Cohn}]{yang2012detecting}
\bibinfo{author}{Y.~Yang}, \bibinfo{author}{C.~Fairbairn}, \bibinfo{author}{J.~F. Cohn},
\newblock \bibinfo{title}{Detecting depression severity from vocal prosody},
\newblock \bibinfo{journal}{IEEE transactions on affective computing} \bibinfo{volume}{4} (\bibinfo{year}{2012}) \bibinfo{pages}{142--150}.
\bibitem[{DeVault et~al.(2014)DeVault, Artstein, Benn, Dey, Fast, Gainer, Georgila, Gratch, Hartholt, Lhommet et~al.}]{devault2014simsensei}
\bibinfo{author}{D.~DeVault}, \bibinfo{author}{R.~Artstein}, \bibinfo{author}{G.~Benn}, \bibinfo{author}{T.~Dey}, \bibinfo{author}{E.~Fast}, \bibinfo{author}{A.~Gainer}, \bibinfo{author}{K.~Georgila}, \bibinfo{author}{J.~Gratch}, \bibinfo{author}{A.~Hartholt}, \bibinfo{author}{M.~Lhommet}, et~al.,
\newblock \bibinfo{title}{Simsensei kiosk: A virtual human interviewer for healthcare decision support},
\newblock in: \bibinfo{booktitle}{Proceedings of the 2014 international conference on Autonomous agents and multi-agent systems}, \bibinfo{year}{2014}, pp. \bibinfo{pages}{1061--1068}.
\bibitem[{Ambadar et~al.(2009)Ambadar, Cohn, and Reed}]{ambadar2009all}
\bibinfo{author}{Z.~Ambadar}, \bibinfo{author}{J.~F. Cohn}, \bibinfo{author}{L.~I. Reed},
\newblock \bibinfo{title}{All smiles are not created equal: Morphology and timing of smiles perceived as amused, polite, and embarrassed/nervous},
\newblock \bibinfo{journal}{Journal of nonverbal behavior} \bibinfo{volume}{33} (\bibinfo{year}{2009}) \bibinfo{pages}{17--34}.
\bibitem[{Utami and Bickmore(2019)}]{utami2019collaborative}
\bibinfo{author}{D.~Utami}, \bibinfo{author}{T.~Bickmore},
\newblock \bibinfo{title}{Collaborative user responses in multiparty interaction with a couples counselor robot},
\newblock in: \bibinfo{booktitle}{2019 14th ACM/IEEE International Conference on Human-Robot Interaction (HRI)}, \bibinfo{organization}{IEEE}, \bibinfo{year}{2019}, pp. \bibinfo{pages}{294--303}.
\bibitem[{Ward and Tsukahara(2000)}]{ward2000prosodic}
\bibinfo{author}{N.~Ward}, \bibinfo{author}{W.~Tsukahara},
\newblock \bibinfo{title}{Prosodic features which cue back-channel responses in english and japanese},
\newblock \bibinfo{journal}{Journal of pragmatics} \bibinfo{volume}{32} (\bibinfo{year}{2000}) \bibinfo{pages}{1177--1207}.
\bibitem[{Benus et~al.(2007)Benus, Gravano, and Hirschberg}]{benus2007prosody}
\bibinfo{author}{S.~Benus}, \bibinfo{author}{A.~Gravano}, \bibinfo{author}{J.~B. Hirschberg},
\newblock \bibinfo{title}{The prosody of backchannels in american english}  (\bibinfo{year}{2007}).
\bibitem[{Bertrand et~al.(2007)Bertrand, Ferr{\'e}, Blache, Espesser, and Rauzy}]{bertrand2007backchannels}
\bibinfo{author}{R.~Bertrand}, \bibinfo{author}{G.~Ferr{\'e}}, \bibinfo{author}{P.~Blache}, \bibinfo{author}{R.~Espesser}, \bibinfo{author}{S.~Rauzy},
\newblock \bibinfo{title}{Backchannels revisited from a multimodal perspective},
\newblock in: \bibinfo{booktitle}{Auditory-visual Speech Processing}, \bibinfo{year}{2007}, pp. \bibinfo{pages}{1--5}.
\bibitem[{Truong et~al.(2011)Truong, Poppe, de~Kok, and Heylen}]{truong2011multimodal}
\bibinfo{author}{K.~P. Truong}, \bibinfo{author}{R.~Poppe}, \bibinfo{author}{I.~de~Kok}, \bibinfo{author}{D.~Heylen},
\newblock \bibinfo{title}{A multimodal analysis of vocal and visual backchannels in spontaneous dialogs.},
\newblock in: \bibinfo{booktitle}{INTERSPEECH}, \bibinfo{year}{2011}, pp. \bibinfo{pages}{2973--2976}.
\bibitem[{Gravano and Hirschberg(2009)}]{gravano2009backchannel}
\bibinfo{author}{A.~Gravano}, \bibinfo{author}{J.~Hirschberg},
\newblock \bibinfo{title}{Backchannel-inviting cues in task-oriented dialogue},
\newblock in: \bibinfo{booktitle}{Tenth Annual Conference of the International Speech Communication Association}, \bibinfo{year}{2009}.
\bibitem[{Wang et~al.(2018)Wang, Alameda-Pineda, Xu, Fua, Ricci, and Sebe}]{wang2018every}
\bibinfo{author}{W.~Wang}, \bibinfo{author}{X.~Alameda-Pineda}, \bibinfo{author}{D.~Xu}, \bibinfo{author}{P.~Fua}, \bibinfo{author}{E.~Ricci}, \bibinfo{author}{N.~Sebe},
\newblock \bibinfo{title}{Every smile is unique: Landmark-guided diverse smile generation},
\newblock in: \bibinfo{booktitle}{Proceedings of the IEEE Conference on Computer Vision and Pattern Recognition}, \bibinfo{year}{2018}, pp. \bibinfo{pages}{7083--7092}.
\bibitem[{Feng et~al.(2017)Feng, Kannan, Gkioxari, and Zitnick}]{feng2017learn2smile}
\bibinfo{author}{W.~Feng}, \bibinfo{author}{A.~Kannan}, \bibinfo{author}{G.~Gkioxari}, \bibinfo{author}{C.~L. Zitnick},
\newblock \bibinfo{title}{Learn2smile: Learning non-verbal interaction through observation},
\newblock in: \bibinfo{booktitle}{2017 IEEE/RSJ International Conference on Intelligent Robots and Systems (IROS)}, \bibinfo{organization}{IEEE}, \bibinfo{year}{2017}, pp. \bibinfo{pages}{4131--4138}.
\bibitem[{Ng et~al.(2022)Ng, Joo, Hu, Li, Darrell, Kanazawa, and Ginosar}]{ng2022learning}
\bibinfo{author}{E.~Ng}, \bibinfo{author}{H.~Joo}, \bibinfo{author}{L.~Hu}, \bibinfo{author}{H.~Li}, \bibinfo{author}{T.~Darrell}, \bibinfo{author}{A.~Kanazawa}, \bibinfo{author}{S.~Ginosar},
\newblock \bibinfo{title}{Learning to listen: Modeling non-deterministic dyadic facial motion},
\newblock in: \bibinfo{booktitle}{Proceedings of the IEEE/CVF Conference on Computer Vision and Pattern Recognition}, \bibinfo{year}{2022}, pp. \bibinfo{pages}{20395--20405}.
\bibitem[{Geng et~al.(2023)Geng, Teotia, Tendulkar, Menon, and Vondrick}]{geng2023affective}
\bibinfo{author}{S.~Geng}, \bibinfo{author}{R.~Teotia}, \bibinfo{author}{P.~Tendulkar}, \bibinfo{author}{S.~Menon}, \bibinfo{author}{C.~Vondrick},
\newblock \bibinfo{title}{Affective faces for goal-driven dyadic communication},
\newblock \bibinfo{journal}{arXiv preprint arXiv:2301.10939}  (\bibinfo{year}{2023}).
\bibitem[{Ertugrul et~al.(2019)Ertugrul, Jeni, Ding, and Cohn}]{ertugrul2019afar}
\bibinfo{author}{I.~O. Ertugrul}, \bibinfo{author}{L.~A. Jeni}, \bibinfo{author}{W.~Ding}, \bibinfo{author}{J.~F. Cohn},
\newblock \bibinfo{title}{Afar: A deep learning based tool for automated facial affect recognition},
\newblock in: \bibinfo{booktitle}{2019 14th IEEE international conference on automatic face \& gesture recognition (FG 2019)}, \bibinfo{organization}{IEEE}, \bibinfo{year}{2019}, pp. \bibinfo{pages}{1--1}.
\bibitem[{Schneider et~al.(2019)Schneider, Baevski, Collobert, and Auli}]{schneider2019wav2vec}
\bibinfo{author}{S.~Schneider}, \bibinfo{author}{A.~Baevski}, \bibinfo{author}{R.~Collobert}, \bibinfo{author}{M.~Auli},
\newblock \bibinfo{title}{wav2vec: Unsupervised pre-training for speech recognition},
\newblock \bibinfo{journal}{arXiv preprint arXiv:1904.05862}  (\bibinfo{year}{2019}).
\bibitem[{McAuliffe et~al.(2017)McAuliffe, Socolof, Mihuc, Wagner, and Sonderegger}]{mcauliffe17_interspeech}
\bibinfo{author}{M.~McAuliffe}, \bibinfo{author}{M.~Socolof}, \bibinfo{author}{S.~Mihuc}, \bibinfo{author}{M.~Wagner}, \bibinfo{author}{M.~Sonderegger},
\newblock \bibinfo{title}{{Montreal Forced Aligner: Trainable Text-Speech Alignment Using Kaldi}},
\newblock in: \bibinfo{booktitle}{Proc. Interspeech 2017}, \bibinfo{year}{2017}, pp. \bibinfo{pages}{498--502}. \DOIprefix\doi{10.21437/Interspeech.2017-1386}.
\bibitem[{Memon(2020)}]{memon2020acoustic}
\bibinfo{author}{S.~A. Memon},
\newblock \bibinfo{title}{Acoustic correlates of the voice qualifiers: A survey},
\newblock \bibinfo{journal}{arXiv preprint arXiv:2010.15869}  (\bibinfo{year}{2020}).
\bibitem[{Eyben et~al.(2010)Eyben, W{\"o}llmer, and Schuller}]{eyben2010opensmile}
\bibinfo{author}{F.~Eyben}, \bibinfo{author}{M.~W{\"o}llmer}, \bibinfo{author}{B.~Schuller},
\newblock \bibinfo{title}{Opensmile: the munich versatile and fast open-source audio feature extractor},
\newblock in: \bibinfo{booktitle}{Proceedings of the 18th ACM international conference on Multimedia}, \bibinfo{year}{2010}, pp. \bibinfo{pages}{1459--1462}.
\bibitem[{Pennebaker et~al.(2015)Pennebaker, Boyd, Jordan, and Blackburn}]{pennebaker2015development}
\bibinfo{author}{J.~W. Pennebaker}, \bibinfo{author}{R.~L. Boyd}, \bibinfo{author}{K.~Jordan}, \bibinfo{author}{K.~Blackburn}, \bibinfo{title}{The development and psychometric properties of LIWC2015}, \bibinfo{type}{Technical Report}, \bibinfo{year}{2015}.
\bibitem[{Ekstedt and Skantze(2022)}]{ekstedt2022voice}
\bibinfo{author}{E.~Ekstedt}, \bibinfo{author}{G.~Skantze},
\newblock \bibinfo{title}{Voice activity projection: Self-supervised learning of turn-taking events},
\newblock \bibinfo{journal}{arXiv preprint arXiv:2205.09812}  (\bibinfo{year}{2022}).
\bibitem[{Hershey et~al.(2017)Hershey, Chaudhuri, Ellis, Gemmeke, Jansen, Moore, Plakal, Platt, Saurous, Seybold et~al.}]{hershey2017cnn}
\bibinfo{author}{S.~Hershey}, \bibinfo{author}{S.~Chaudhuri}, \bibinfo{author}{D.~P. Ellis}, \bibinfo{author}{J.~F. Gemmeke}, \bibinfo{author}{A.~Jansen}, \bibinfo{author}{R.~C. Moore}, \bibinfo{author}{M.~Plakal}, \bibinfo{author}{D.~Platt}, \bibinfo{author}{R.~A. Saurous}, \bibinfo{author}{B.~Seybold}, et~al.,
\newblock \bibinfo{title}{Cnn architectures for large-scale audio classification},
\newblock in: \bibinfo{booktitle}{2017 ieee international conference on acoustics, speech and signal processing (icassp)}, \bibinfo{organization}{IEEE}, \bibinfo{year}{2017}, pp. \bibinfo{pages}{131--135}.
\bibitem[{Bahdanau et~al.(2014)Bahdanau, Cho, and Bengio}]{bahdanau2014neural}
\bibinfo{author}{D.~Bahdanau}, \bibinfo{author}{K.~Cho}, \bibinfo{author}{Y.~Bengio},
\newblock \bibinfo{title}{Neural machine translation by jointly learning to align and translate},
\newblock \bibinfo{journal}{arXiv preprint arXiv:1409.0473}  (\bibinfo{year}{2014}).
\bibitem[{Stoll et~al.(2018)Stoll, Camg{\"o}z, Hadfield, and Bowden}]{stoll2018sign}
\bibinfo{author}{S.~Stoll}, \bibinfo{author}{N.~C. Camg{\"o}z}, \bibinfo{author}{S.~Hadfield}, \bibinfo{author}{R.~Bowden},
\newblock \bibinfo{title}{Sign language production using neural machine translation and generative adversarial networks},
\newblock in: \bibinfo{booktitle}{Proceedings of the 29th British Machine Vision Conference (BMVC 2018)}, \bibinfo{organization}{British Machine Vision Association}, \bibinfo{year}{2018}.
\bibitem[{Ahuja et~al.(2020)Ahuja, Lee, Ishii, and Morency}]{ahuja2020no}
\bibinfo{author}{C.~Ahuja}, \bibinfo{author}{D.~W. Lee}, \bibinfo{author}{R.~Ishii}, \bibinfo{author}{L.-P. Morency},
\newblock \bibinfo{title}{No gestures left behind: Learning relationships between spoken language and freeform gestures},
\newblock in: \bibinfo{booktitle}{Findings of the Association for Computational Linguistics: EMNLP 2020}, \bibinfo{year}{2020}, pp. \bibinfo{pages}{1884--1895}.
\bibitem[{Al~Moubayed et~al.(2012)Al~Moubayed, Beskow, Skantze, and Granstr{\"o}m}]{al2012furhat}
\bibinfo{author}{S.~Al~Moubayed}, \bibinfo{author}{J.~Beskow}, \bibinfo{author}{G.~Skantze}, \bibinfo{author}{B.~Granstr{\"o}m},
\newblock \bibinfo{title}{Furhat: a back-projected human-like robot head for multiparty human-machine interaction},
\newblock in: \bibinfo{booktitle}{Cognitive Behavioural Systems: COST 2102 International Training School, Dresden, Germany, February 21-26, 2011, Revised Selected Papers}, \bibinfo{organization}{Springer}, \bibinfo{year}{2012}, pp. \bibinfo{pages}{114--130}.

\end{thebibliography}

\newpage

\section{Appendix}
\subsection{Distribution of Intensity and Duration of Smiles}
\begin{figure}[!h]
    \centering
    \includegraphics[width=\linewidth]{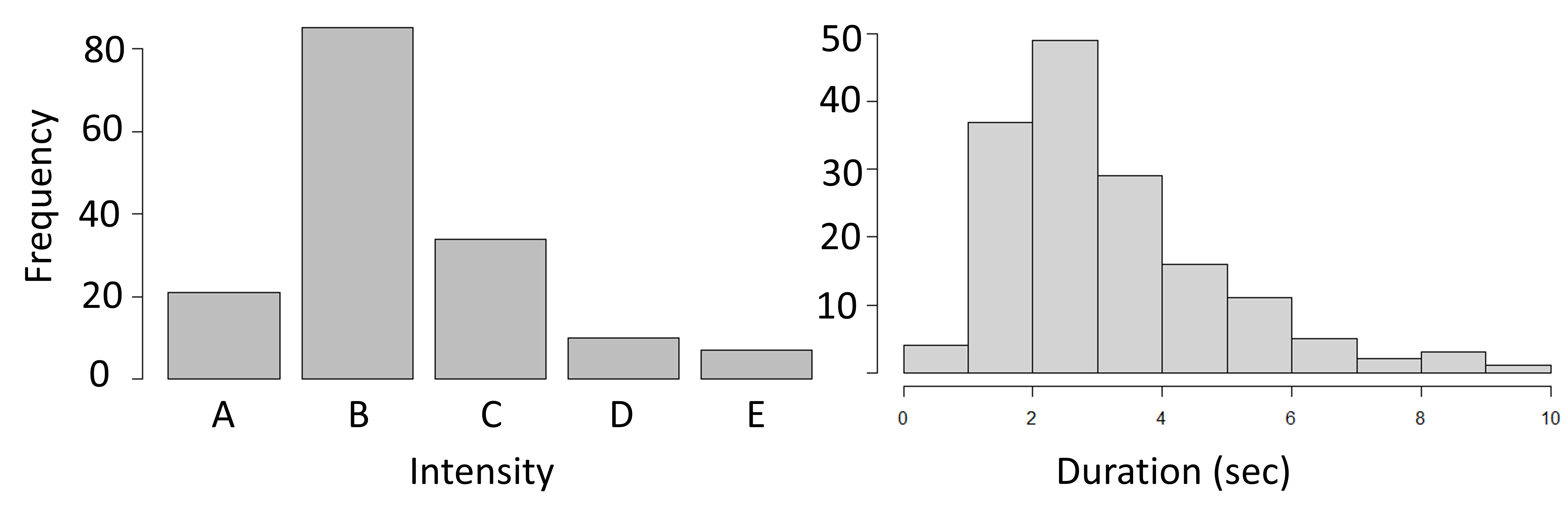}
    \caption{Distribution of intensity and duration of BC smiles in the annotated dataset. The spread of the histograms shows the diversity of the annotated smiles.}
    \label{fig:intensity_duration_distribution}
\end{figure}

Figure \ref{fig:intensity_duration_distribution} shows the distribution of annotated Backchannel (BC) smiles in terms of their intensity and duration. The predicted intensity using the automated approach showed that over 50\% of smiles were of B-level intensity, and fewer instances of high-intensity smiles (D and E-levels) were also present. The mean duration was 3.18 $\pm$ 1.71 seconds.

\subsection{Effect of Sex and Relationship on Smile Intensity}

\begin{table}[h]
\caption{ANOVA of listener sex, speaker sex, and relationship on intensity of smile. `{\dotr}' indicates significant at p\textless0.1.}
\centering
\resizebox{\columnwidth}{!}{
\begin{tabular}{lrrrrr}
  \toprule
 & Df & Sum Sq & Mean Sq & F value & Pr($>$F) \\ 
  \midrule
$sex_{listener}$ & 1 & 0.53 & 0.53 & 0.60 & 0.4417 \\ 
$sex_{speaker}$ & 1 & 2.93 & 2.93 & 3.31 & 0.0710 ${\dotr}$ \\ 
$relationship$ & 3 & 3.23 & 1.08 & 1.22 & 0.3055 \\ 
\begin{tabular}{c}
    $sex_{listener}*$ \\ 
    $relationship$
\end{tabular} & 3 & 2.00 & 0.67 & 0.75 & 0.5225 \\ 
\begin{tabular}{c}
     $sex_{listener}*$ \\
     $sex_{speaker}$ 
\end{tabular} & 1 & 0.10 & 0.10 & 0.11 & 0.7424 \\ 
\begin{tabular}{c}
     $sex_{speaker}*$ \\
     $relationship$
\end{tabular} & 3 & 3.15 & 1.05 & 1.19 & 0.3176 \\ 
\midrule
  Residuals & 144 & 127.49 & 0.89 &  &  \\ 
   \bottomrule
\end{tabular}
}
\label{anova_intensity}
\end{table}


Note that the intensity of the smile differs marginally by the speaker sex. It is not affected by other factors such as relationship, listener sex and their interaction.

\end{document}